\documentclass[a4paper,conference]{IEEEtran}
\usepackage{times}
\usepackage{epsfig}
\usepackage{graphicx}
\usepackage{caption}
\usepackage{subcaption}
\usepackage{amsmath}
\usepackage{amssymb}
\usepackage{algorithm}
\usepackage{algorithmic}
\newcommand{\rmnum}[1]{\romannumeral #1}
\newcommand{\Rmnum}[1]{\expandafter\@slowromancap\romannumeral #1@}
\ifCLASSINFOpdf
\else
\fi
\begin{document}
\title{Salient Region Segmentation}

\author{
	\IEEEauthorblockN{
		Sen He\IEEEauthorrefmark{1},
		Nicolas Pugeault\IEEEauthorrefmark{2}
	}
	\IEEEauthorblockA{
		Department of Computer Science\\
		University of Exeter\\
		Harrison Building, Streatham Campus,
		Exeter, EX4 4QF\\
		Email: 
		\IEEEauthorrefmark{1}sh752@exeter.ac.uk,
		\IEEEauthorrefmark{2}N.Pugeault@exeter.ac.uk
	}
}
\maketitle

\begin{abstract}
Saliency prediction is a well studied problem in computer vision. Early saliency models were based on low-level hand-crafted feature derived from insights gained in neuroscience and psychophysics. In the wake of deep learning breakthrough, a new cohort of models were proposed based on neural network architectures, allowing significantly higher gaze prediction than previous shallow models, on all metrics. 
However, most models treat the saliency prediction as a \textit{regression} problem, and accurate regression of high-dimensional data is known to be a hard problem. 
Furthermore, it is unclear that intermediate levels of saliency (ie, neither very high, nor very low) are meaningful: Something is either salient, or it is not.  
Drawing from those two observations, we reformulate the saliency prediction problem as a salient region \textit{segmentation} problem. We demonstrate that the reformulation allows for faster convergence than the classical regression problem, while performance is comparable to state-of-the-art. 
We also visualise the general features learned by the model, which are showed to be consistent with insights from psychophysics.
\end{abstract}

\section{Introduction}\label{sec:intro}
The human visual system receives about $10^{8}$ to $10^{9}$ bits of information per second \cite{Borji2013b}. In order to process such a large quantity of information  efficiently, the visual system relies on dynamic attention: selectively focusing cognitive resources on parts of the scene. This process has been extensively studied by psychologists, some have proposed computational models to mimic this clever mechanism and predict where a human witness would look in a given. The most widespread model is based on Triesman's feature integration theory \cite{Treisman1980} and the concept of centre-surround difference: regions which have features different from their surroundings are likely to attract a viewer's attention. Such regions, which attract a viewer's gaze, are called \textit{salient}, and thus computational models predicting such regions are called \textit{saliency models}.
Features commonly used to explain saliency include, eg, colour, intensity and orientation \cite{Itti2001}, and additional features (eg, depth, contour) can improve the saliency models' predictiveness but it is not fully understood which features are salient. The fast development and success of deep learning approaches in computer vision, together with the availability of large scale datasets in visual attention has allowed to try and learn salient features from scratch by trying to predict human viewers' gaze \cite{kummerer2014deep,Kummerer2016,Jetley2016a,kruthiventi2015deepfix}. These data-driven approaches have demonstrated the potential to predict viewers' gaze with significantly higher accuracy than previous, hand-crafted saliency models. 

One challenge when using a data-driven approach to learn saliency from human gaze, is the difficulty of regressing a complete, high-dimensional saliency map: accurate regression of high-dimensional data is known to be a hard problem. 
More importantly, although it is clear what is signified by a high or low saliency, it is less clear how meaningful is the accurate saliency scoring of intermediate regions (can something be `somewhat more salient' than another?).  
Arguably, the regression problem to be solved is much harder than strictly necessary from the problem definition: estimating accurately the relative saliency level of all pixels is intrinsically ambiguous, whereas we are only really interested in the salient regions of the image. 
In this work, instead of predicting a pixel-wise saliency value, we predict salient \textit{regions} based on three saliency levels (high, medium and low). This simplification of the problem allows us to adopt an image segmentation approach based on an encoder-decoder network such that the network's output is of the same size as its input, avoiding the need to rescale the output.
We demonstrate that although our approach involves a major reformulation and simplification of the problem, we can predict viewers' gaze with an accuracy comparable to state-of-the-art approaches. 

\begin{figure}%[h!]
	\begin{center}
		\begin{subfigure}{0.24\textwidth}
			\includegraphics[width=\textwidth]{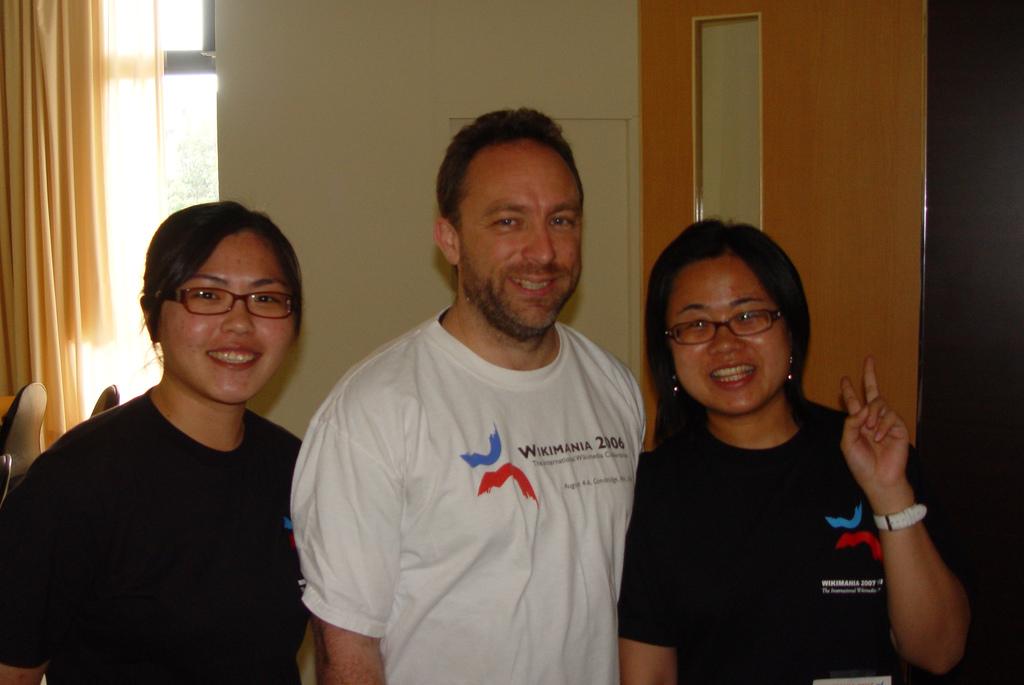}
			\caption{Original image}
		\end{subfigure}
		\begin{subfigure}{0.24\textwidth}
			\includegraphics[width=\textwidth]{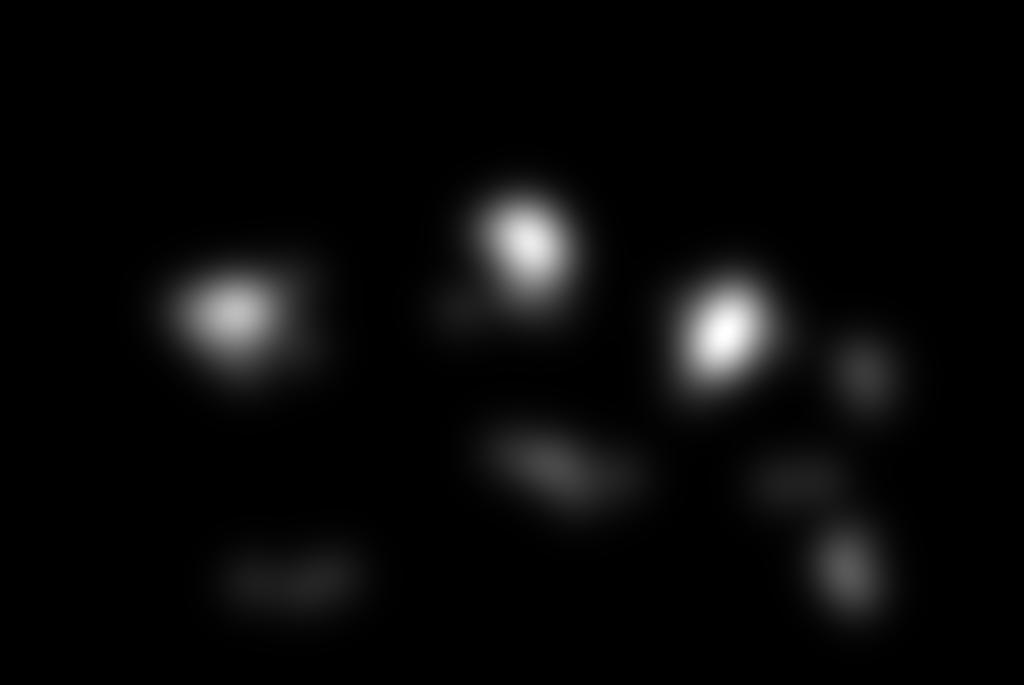}
			\caption{Saliency Map (SM)}
		\end{subfigure}
		\begin{subfigure}{0.24\textwidth}
			\includegraphics[width=\textwidth]{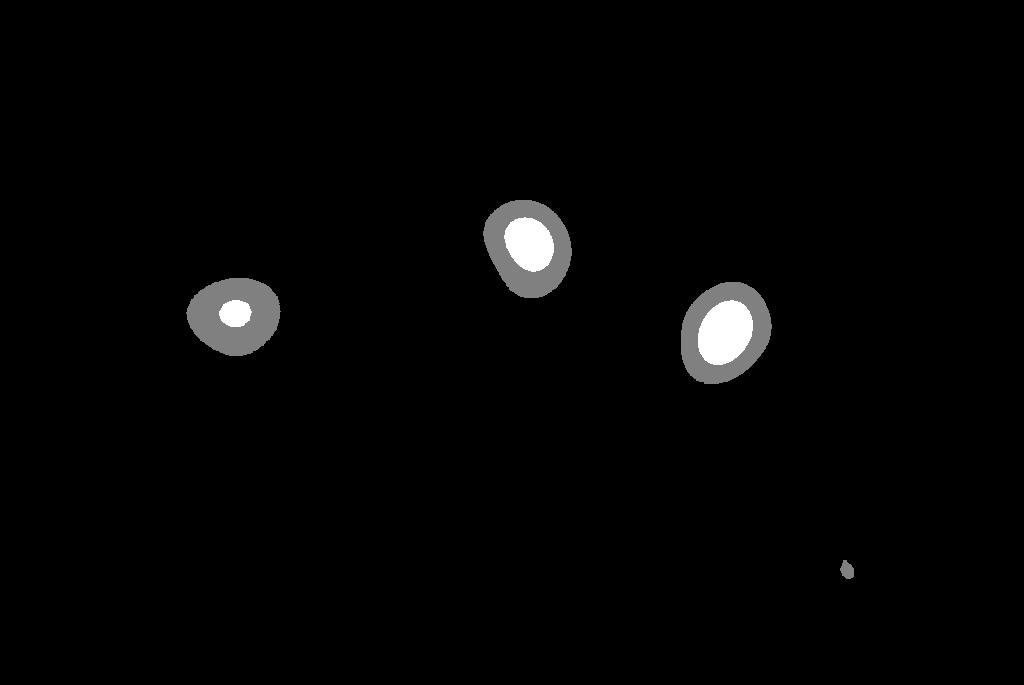}
			\caption{Salient Region Map (SRM)}
		\end{subfigure}
	\end{center}
	\caption{
		Example of original image, saliency map and salient region map (three saliency levels).
	}
	\label{fig:saliencyregion}
\end{figure}

\section{Related work}
Existing saliency prediction methods can be divided into two groups based on whether the features used in saliency prediction are hand-crafted or learned from the data. We will discuss those two categories in turn.

\subsection{Models based on hand-crafted features}
Most hand-crafted feature-based models originate from the Treisman's feature integration theory~\cite{Treisman1980}, and make use of hand-crafted features based on what has been shown to attract visual attention by Psychologists. 
Itti \& Koch~\cite{Itti2000a,Itti2001} were the first to propose a computational model for saliency prediction based on the feature integration theory and central surround difference. Their model uses a topological architecture at multiple scale to get a conspicuity map for each feature, and then use a combination scheme to combine conspicuity maps for all the features used in saliency prediction to get the final saliency map. Their model uses three common features: colour, intensity and orientation. Later variants on this model~\cite{Ouerhani2000} also integrate the depth feature into the saliency prediction model. However, many other features could attract our attention; for detailed survey of hand-crafted feature based saliency prediction model, we refer to~\cite{Borji2013b}.

\subsection{Models based on machine learning method}
Although psychological research has provided elements of answer, what visual features are salient is still an open question, hence there is an interest to use machine learning to learn salient features from data. 
Kienzle et al.~\cite{Kienzle2009a} first tried to learn features from eye fixation data, and found confirmation that visual patterns similar to central surround difference attract viewers' attention. Since 2014, the development of deep learning approaches and their successes in many computer vision problems, along with the availability of large scale datasets in visual attention, lead to the rise of deep learning approaches in saliency prediction. 

Vig et al.~\cite{vig2014large} were the first to use a deep neural network for saliency prediction, although many other approaches followed. They use an optimisation algorithm to search the best features in a deep model created for face recognition. After that, most deep models for saliency prediction use transfer learning method and fine-tuning on a deep network pre-trained for image classification with some modification, removing the fully connected layer and building a regressor above the convolutional part of the network. The final step is to rescale the network's output to the input image's size. The main difference between those deep models is the loss function used for training: \cite{Kruthiventi2015} directly use Euclidean loss function to train their model; \cite{Corniaa} use the Kullback-Leibler divergence as the loss function, as it is one of the evaluation metric for saliency model; \cite{Jetley2016a} compares different loss functions and found that the Bhattacharyya distance is the best loss function; and  \cite{Kummerer2016} use the maximum likelihood method to train their model.

The approach proposed in this article is also based on deep learning, but in contrast to those who treat saliency prediction as a regression problem, we reformulate the problem as a segmentation problem, and we use the encoder-decoder architecture to achieve pixel-wise prediction rather than the coarse prediction provided by other deep networks.

\begin{figure*}%[h!]
	\begin{center}
		\begin{subfigure}{0.3\textwidth}
			\includegraphics[width=\textwidth]{original}
			\caption{input}
		\end{subfigure}
		\begin{subfigure}{0.3\textwidth}
			\includegraphics[width=\textwidth]{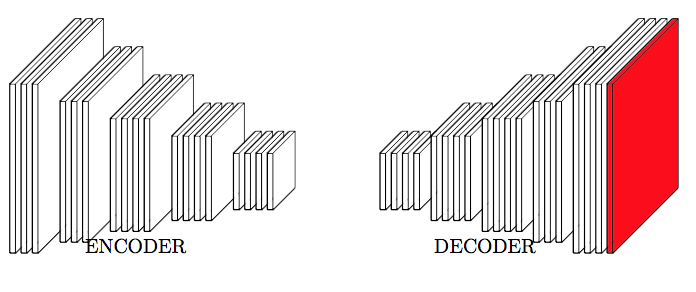}
			\caption{network}
		\end{subfigure}
		\begin{subfigure}{0.3\textwidth}
			\includegraphics[width=\textwidth]{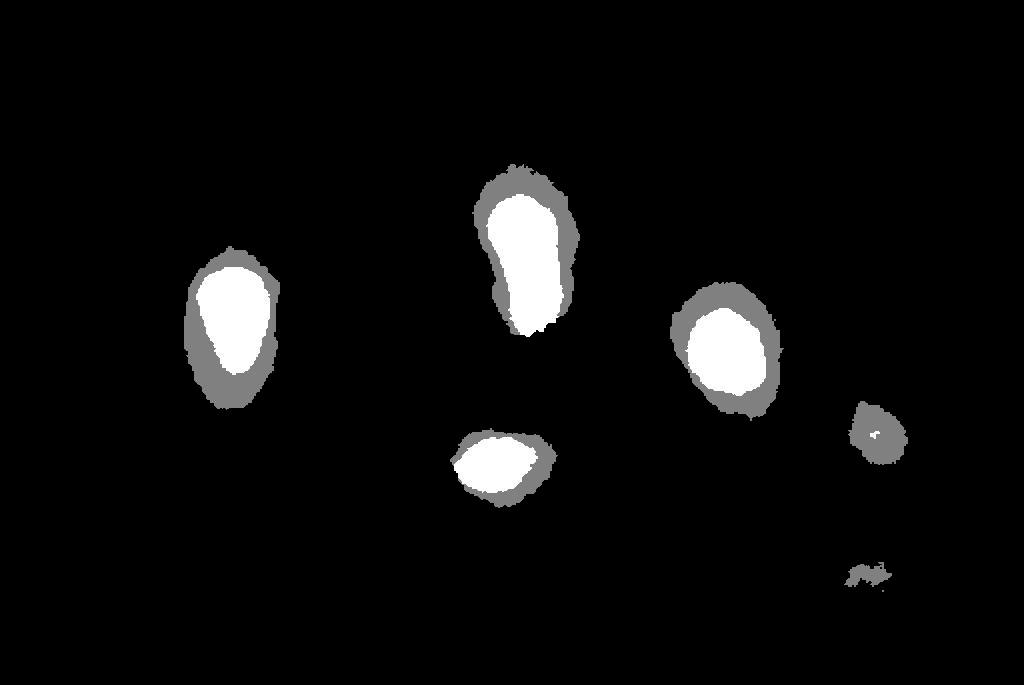}
			\caption{output}
		\end{subfigure}
	\end{center}
	\caption{Model architecture: this is a fully convolutional network without fully connected layer. The last red layer is the softmax with cross entropy loss layer in the training stage (it is replaced by the argmax function in the inference stage).}
	\label{fig:model}
\end{figure*}

\section{Salient Region Segmentation}
Before introducing our method, we first introduce three terms that will be used in the later parts (example in Figure~\ref{fig:saliencyregion}).

\textbf{Fixation map (FM):} A fixation map is a binary map, which records the human eye fixation locations using eye tracker when generating the dataset for saliency prediction.

\textbf{Saliency map (SM):} A saliency map is derived from fixation maps by convolving a Gaussian filter with the fixation map. It is the ground truth for all of the saliency prediction model at the moment.

\textbf{Salient region map (SRM):} A salient region map is derived by quantizing a saliency map into several saliency levels. It is the ground truth in our method. See Figure~\ref{fig:saliencyregion}.

All existing saliency prediction models work by trying to predict saliency levels at every pixel. However, estimating intermediate saliency values accurately increases significantly the complexity of the learning problem, with little benefit. 
Therefore, we propose to reformulate the problem as a segmentation problem to separate regions of high saliency from regions of  low saliency, rather than regressing a pixel-wise saliency map.

\subsection{Salient Region Thresholding}
The saliency map is a continuous map while the salient region map is a discrete map. The salient region map is derived directly from the saliency map by assigning each pixel a saliency level according to their saliency value. The higher the saliency value of a pixel, the higher its saliency level:
\begin{align}
R(x)=
\begin{cases}
0 & \text{if } 0 \leqslant S(x) < \frac{255}{K}\times1
\\
\frac{255}{K-1} & \text{if } \frac{255}{K}\times1 \leqslant S(x) < \frac{255}{K}\times2
\\
\vdots & \vdots
\\
255 & \text{if } frac{255}{K}\times$(K-1)$ \leqslant S(X) < \frac{255}{K}\times K
\end{cases}
\end{align}
where $R(x)$ is the saliency level of salient region map at pixel $x$ , $S(x)$ is the saliency value of saliency map at pixel $x$, and $K$ is the number of saliency levels. The resulting salient region map will be used as ground truth in the training stage.

\subsection{Encoder-Decoder Segmentation}
The classical Convolutional Neural Network is an encoder plus a classifier or a regressor. The size of such a network's output is very small compared to the input image due to the successive pooling layers, therefore preventing pixel-wise classification in this architecture. Recently, the encoder-decoder architecture has achieved great performance in semantic segmentation task; and therefore this is the architecture we adopt for salient region segmentation in this paper. 
There are mainly two different encoder-decoder architectures: the FCN (Fully Convolutional Networks)~\cite{Long2015} and SegNet~\cite{Badrinarayanan2015}. The main difference between these two approaches is the upsampling process used in the decoder. In FCN, they learn an upsampling filter to do deconvolution against the corresponding pooling layer such that the output size is equal to the input size. In contrast, SegNet uses indices in the pooling stage, and do unpooling to get a sparse feature map at first in the decoder (this method is also used in CNN feature visualisation of \cite{zeiler2014visualizing}), then learn a filter to get a dense feature map. As the SegNet architecture has a higher classification accuracy than FCN for semantic segmentation, we adopt this architecture for salient region segmentation. 

\subsection{Median Frequency Balancing} 
For training our network, we use a softmax with cross entropy loss function. However, the dataset is highly unbalanced: low saliency regions occupies a much larger proportion of the whole map than high saliency regions do (see Figure \ref{fig:saliencyregion}). Therefore, we use Eigen and Fergus's median frequency balancing method~\cite{Eigen2014} to weight the loss caused by each class:
\begin{align}
E(x)=\sum_{i=1}^{K}{W_i}\left({g_i}(x)\ln{p_i}(x)+(1-{g_i}(x))\ln(1-{p_i}(x))\right)
\end{align}
where $E(x)$ is the loss caused by pixel $x$ during training, $K$ is the number of saliency levels, ${W_i}$ is the weight of $i^\text{th}$ saliency level determined by median frequency balancing method,  ${p_i}$ is the prediction probability that the saliency level of pixel $x$ is $i$, and ${g_i}$ is the ground truth that the saliency level of pixel $x$ is $i$. 

The whole model architecture is in Figure ~\ref{fig:model}.

\subsection{Region Restriction}\label{regionrestrict}
\begin{figure*}[h!]
	\begin{center}
		\begin{subfigure}{0.3\textwidth}
			\includegraphics[width=\textwidth]{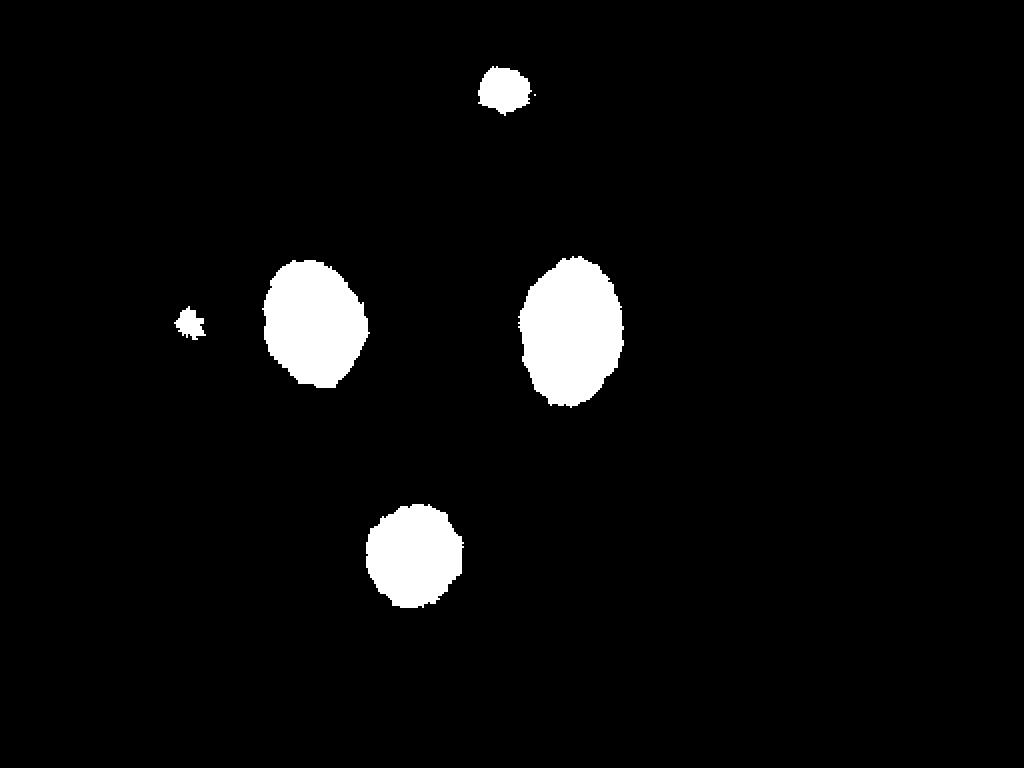}
			\caption{2 levels}
		\end{subfigure}
		\begin{subfigure}{0.3\textwidth}
			\includegraphics[width=\textwidth]{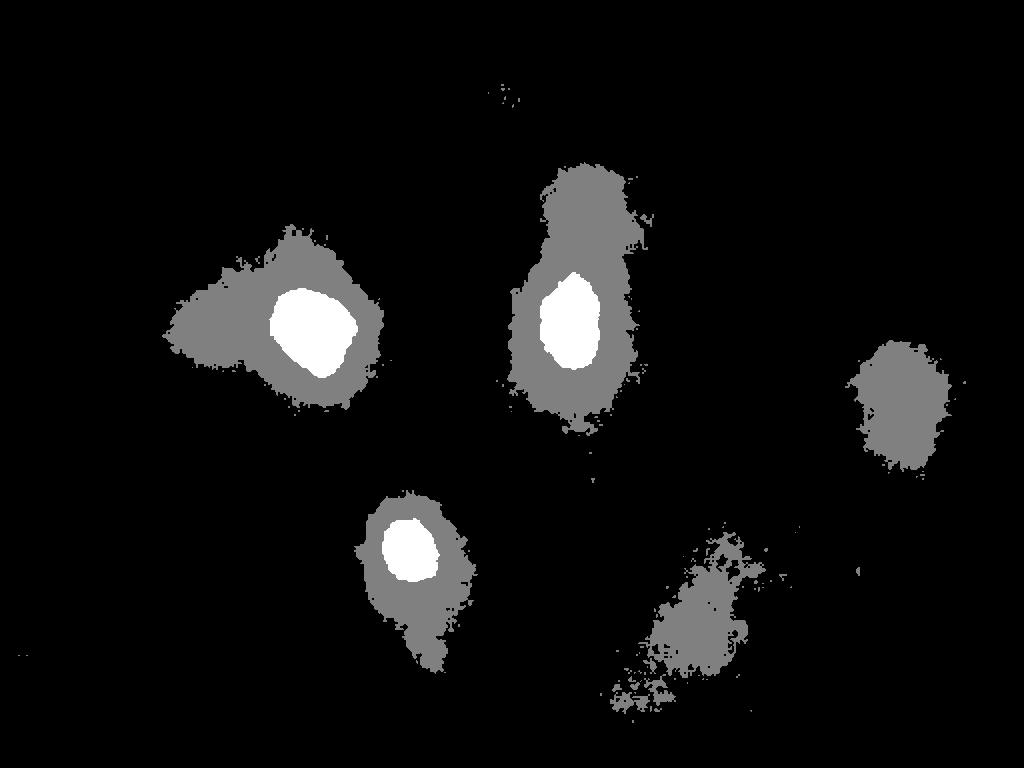}
			\caption{3 levels}
		\end{subfigure}
		\begin{subfigure}{0.3\textwidth}
			\includegraphics[width=\textwidth]{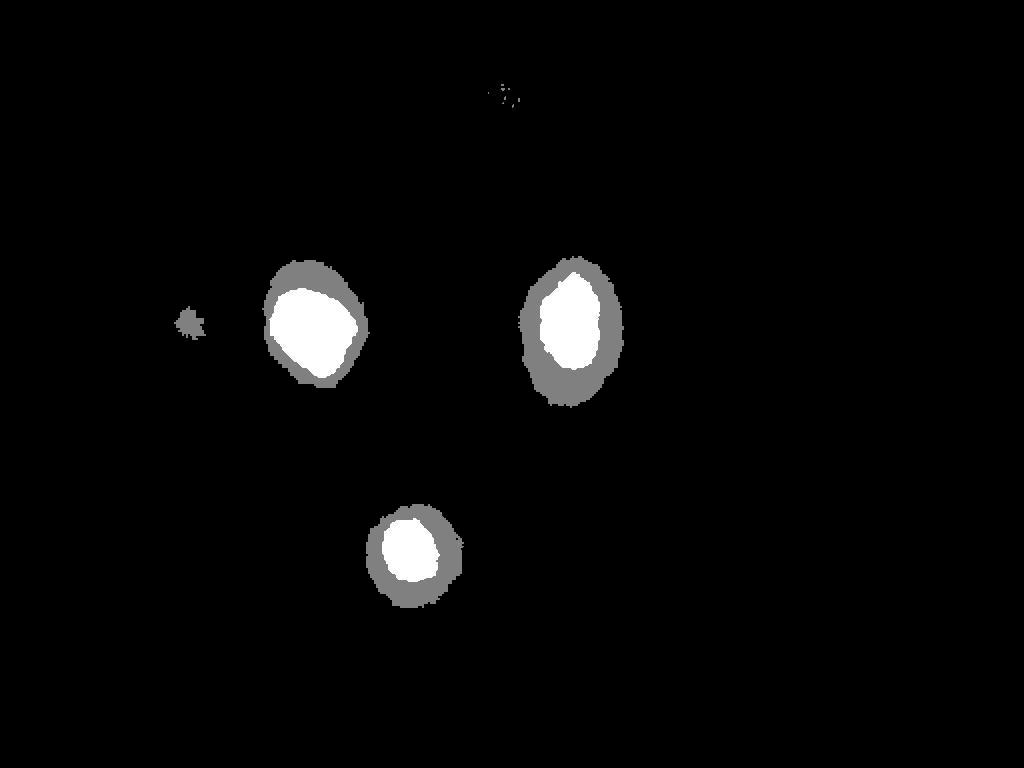}
			\caption{restriction}
		\end{subfigure}
	\end{center}
	\caption{Region Restriction: the output of two saliency levels model, the output of three saliency levels model and the output after region restriction.}
	\label{fig:r_restrict}
\end{figure*}

To make the salient region more selective, we use the output of a binary (salient/non-salient) model to restrict the regions of the output with more quantization levels, such that the pixels that are non-salient in the binary model's output are inhibited in the quantized model's output (example in Figure \ref{fig:r_restrict}).

\subsection{Training The Model}
We trained two models, one model with three saliency levels and one model with two saliency levels. We use the second model's output to restrict the first model's output. The parameters of the encoder part are initialized from the VGG-16 network \cite{simonyan2014very}. The filters in the decoder are initialized by MSRA method in Caffe \cite{Jia2014}. During the training stage, all layers are learnt with an initial learning rate of $0.01$, we decrease the learning rate using the step learning policy with a step size of 500. All training is based on the SALICON dataset \cite{jiang2015salicon}. We only use the training dataset (10K images) in SALICON to train our model, and we use the first 500 images in the validation dataset of SALICON as validation. It takes almost 22 hours to train each model on a K40 GPU.

\section{Evaluation}\label{sec:eval}
In this section, we first describe the datasets and the evaluation metrics for saliency prediction models, followed by a comparison and discussion of the quantitative and qualitative aspects of the result.

\subsection{Datasets}
\textbf{SALICON} \cite{jiang2015salicon}: This is the largest dataset publicly available with fixation map and saliency map in visual attention domain. It consists of 20,000 images taken from the MSCOCO dataset, 10,000 in the training dataset, 5,000 in the validation dataset, and 5,000 in the testing dataset. The eye fixation data is recorded by mouse-click instead of eye-tracking system. We use the training dataset for training the model, and the first 500 data in the validation dataset for validation and all the validation data for testing the classification accuracy.

\textbf{MIT1003} \cite{judd2009learning}: This dataset contains 1,003 images with fixation and saliency map. The fixation map comes from 15 viewers when free viewing the original image for 3 seconds recorded by the eye-tracking system. We use this dataset to evaluate the loss in several saliency evaluation metrics compared to the traditional pixel-wise saliency value prediction methods.

\textbf{MIT300} \cite{judd2012benchmark}: This dataset contains 300 images and it is the MIT saliency benchmark. The fixation map (not publicly available) of this dataset recorded the fixation locations of 39 people. We use this dataset to compare the performance of our model with other state-of-the-art saliency prediction models over several evaluation metrics.

\subsection{Metrics} 
The ground truth for our model is the salient region map, not the saliency map. Therefore, we evaluate our model using the classification accuracy and the evaluation metrics based on the fixation map.

\textbf{Classification accuracy}: This is the evaluation used in semantic segmentation tasks, we test the overall classification accuracy (the accuracy for all saliency levels), per class accuracy (the accuracy for each saliency level).

\textbf{AUC-Judd}: This metric is proposed by Judd in \cite{Riche2013}. The model prediction is treated as a binary classifier to separate positive from negative samples at various thresholds, and the ROC curve is calculated from true positive (TP) and false positive (FP) rates. The final score is the area under the ROC curve, where larger AUC signify better detection. 

\textbf{sAUC}: This metric is introduced in \cite{zhang2008sun} at 2008. It is the same as AUC-Borji but removed the central-bias (when photographing the image, people like to place the interesting objects in the central part of the image).

\textbf{NSS} \cite{Riche2013}: This metric is the normalised scanpath saliency between the model's prediction and the fixation map. It is measured as the mean value of the normalised model prediction at fixation locations.

\subsection{Evaluation results}
\subsubsection{Classification accuracy}
Table~\ref{table:3lvl} records the proposed model's classification accuracy on the SALICON dataset, overall and for each saliency level, 
showing that classification accuracy is lower for higher saliency levels. 
\begin{table}%[h]
\centering
\begin{tabular}{|c|c|}
\hline
Class             & Accuracy\\
\hline
all class         & 84.21\%\\
saliency level 1  & 86.6\%\\
saliency level 2  & 68.11\%\\
saliency level 3  & 65.11\%\\
\hline
\end{tabular}
\caption{The overall classification accuracy (all class in the table), saliency level 1, 2 and 3 classification accuracy for 3 saliency levels model.}
\label{table:3lvl}
\end{table} 
In contrast to classical semantic segmentation, an object can belong to multiple classes in salient region segmentation: the high saliency level region is generally surrounded by a low saliency level region. It makes it difficult to distinguish them with higher accuracy, as they easily overlap (in particular, the low saliency region covers the high saliency region). 

\subsubsection{Saliency evaluation metrics}
Due to the quantization step, it is expected that even a true salient region map would yield a somewhat lower gaze prediction performance on most metrics compared to the original (continuous) saliency map. 
This loss is measured one the MIT1003 dataset and recorded in Table~\ref{table:decline}. In this table, we can also see that the estimated loss is larger for AUC metric than for NSS, supporting our hypothesis that quantized regions encode most saliency information.
We evaluated our model on the SALICON validation dataset (Table~\ref{table:salicon}) and the MIT300 official test set (Table~\ref{table:mit}).\\
\begin{table}%[h!]
\centering
\begin{tabular}{|c|c|c|c|c|}
\hline
Ground truth                          &AUC-Judd     &AUC-shuffled   &NSS\\
\hline
saliency map                          &0.9700                  &0.8899               &4.0543\\
salient region map (3 levels) &0.7830                &0.7507              &3.7385\\
\hline
quantization loss                                   &19.28\%            &15.64\%             &7.79\%\\
\hline
\end{tabular}
\caption{The performance decline from saliency map to the salient region map, this is evaluated on the MIT1003 dataset.}
\label{table:decline}
\end{table}
\begin{table}%[h!]
\centering
\begin{tabular}{|c|c|c|c|c|}
\hline
Models                     &AUC-Judd               &AUC-shuffled   &NSS\\
\hline
SALGAN                  &\*                                 &0.773                  &\textbf{2.589}\\
PDP                         & \textbf{0.880}                               &\textbf{0.783}                 &2.419\\
\hline
ours (no region restriction)                        &0.7485                                  &0.6724                 &1.5923\\
ours (with region restriction)                    &0.7008                                   &0.6463              &1.7950\\
\hline
loss compared to state-of-the-art               &20.36\%                      &12.63\%             &30.67\%\\
\hline
\end{tabular}
\caption{Comparison with the state-of-the-art models, all results are tested on SALICON validation dataset, the decline is computed based on the map after region restriction.}
\label{table:salicon}
\end{table}\\
\begin{table}[h!]
\centering
\begin{tabular}{|c|c|c|c|c|}
\hline
Models                     &AUC-Judd               &AUC-shuffled   &NSS\\
\hline
human baseline        &0.92                                 &0.81                 &3.29\\
Deep Gaze \rmnum{2}  &\textbf{0.88}                                 &0.72                 &1.29\\
Deep Gaze \rmnum{1}  &0.84                                 &0.66                 &1.22\\
SALGAN                  &0.86                                 &0.72                  &2.04\\
PDP                         & 0.85                               &\textbf{0.73}                 &\textbf{2.05}\\
\hline
ours (3 levels)                        &0.76                                   &0.68                 &1.32\\
\hline
loss compared to state-of-the-art                &13.64\%                      &6.85\%             &35.61\%\\
\hline
\end{tabular}
\caption{Comparison with the state-of-the-art models, all results are tested on the MIT300 saliency benchmark based on the map without region restriction \cite{mit-saliency-benchmark}.}
\label{table:mit}
\end{table}

\begin{figure*}
	\begin{center}
		\begin{subfigure}{0.2\textwidth}
			\includegraphics[width=\textwidth]{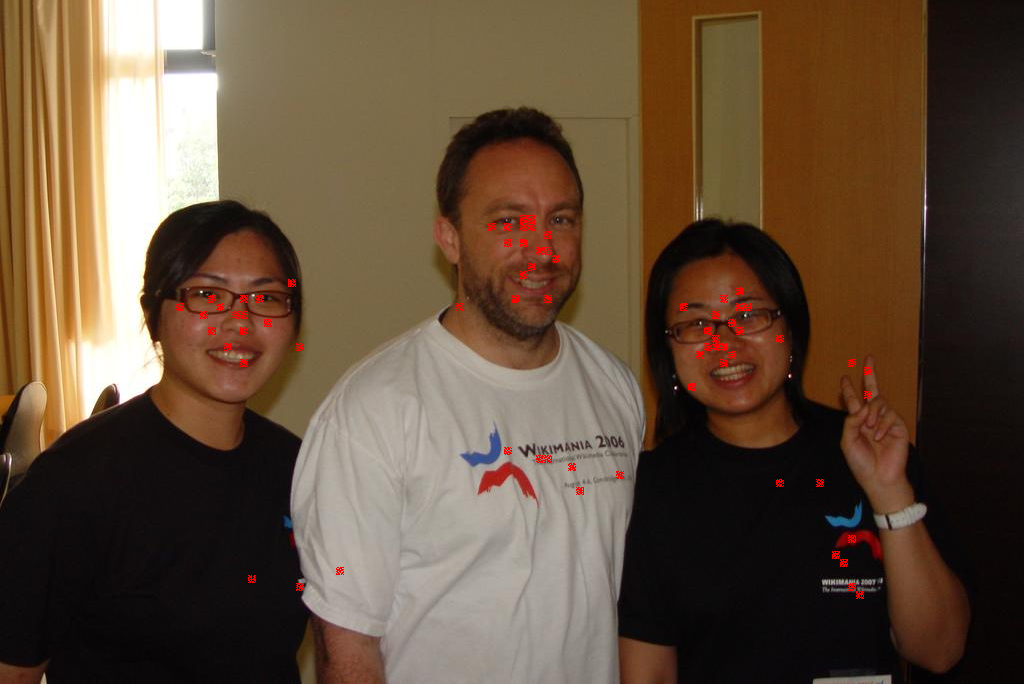}
			\caption{image}
		\end{subfigure}
		\begin{subfigure}{0.2\textwidth}
			\includegraphics[width=\textwidth]{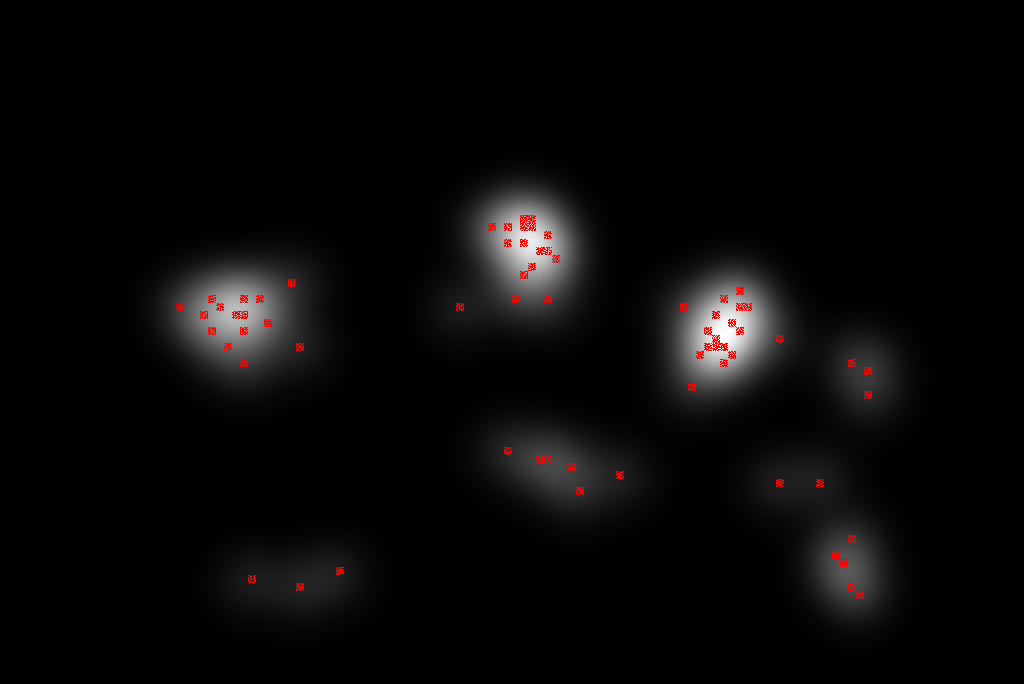}
			\caption{saliency map}
		\end{subfigure}
		\begin{subfigure}{0.2\textwidth}
			\includegraphics[width=\textwidth]{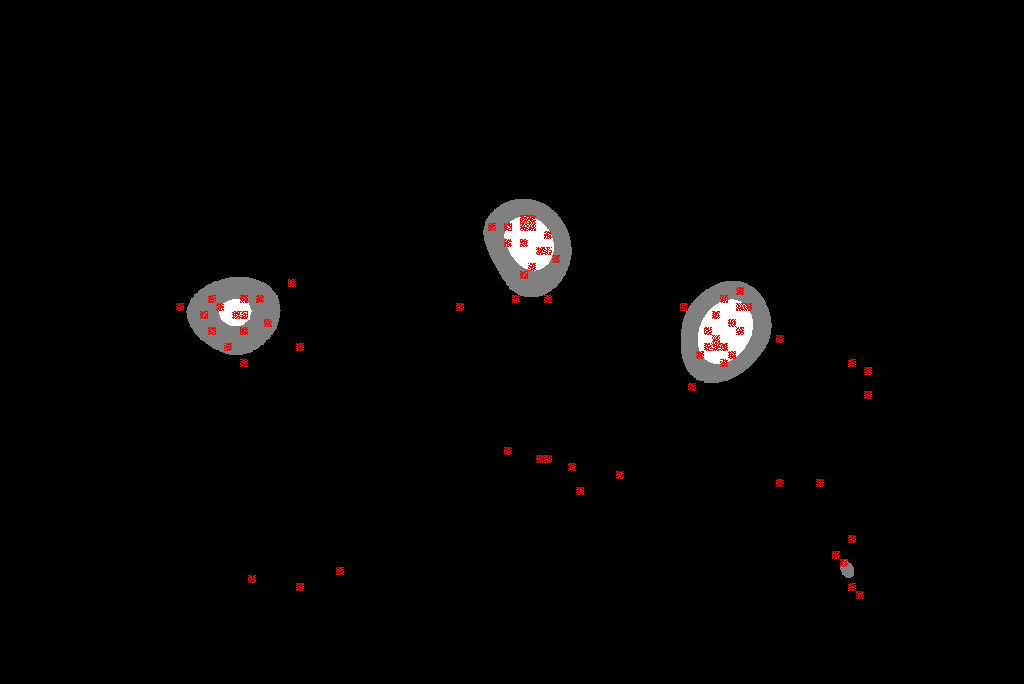}
			\caption{salient region}
		\end{subfigure}
		\begin{subfigure}{0.2\textwidth}
			\includegraphics[width=\textwidth]{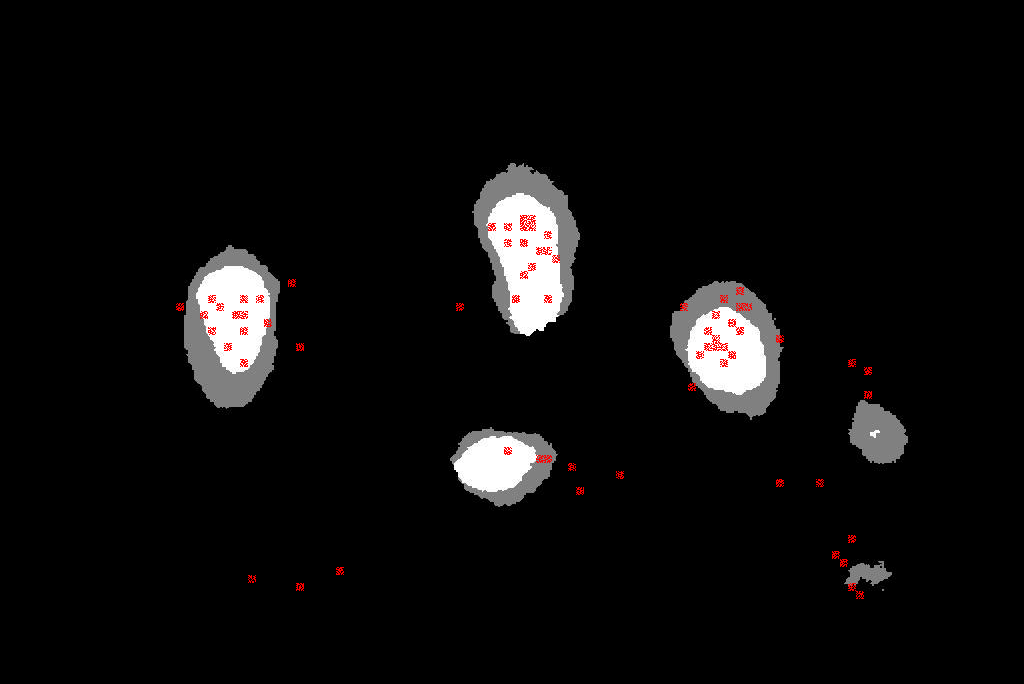}
			\caption{prediction}
		\end{subfigure}
	\end{center}
	\caption{NSS decline explanation, the red spots are human eye fixations recorded when generating the dataset.}
	\label{fig:NSS}
\end{figure*}

These results show that although our proposed model does not quite reach the performance of the latest deep architectures such as SalGAN and PDP, it nonetheless achieves high performance. Indeed, when considering the official MIT300 benchmark (Table~\ref{table:mit}), it is interesting to note that, despite working on quantized regions, our model's NSS score clearly outperforms some previous methods such as Deep Gaze \rmnum{1} , Deep Gaze \rmnum{2}.
Moreover, the results on the SALICON dataset (Table~\ref{table:salicon}) show that the application of the region restriction mechanism described in section \ref{regionrestrict} appears to simultaneously reduce the AUC scores while at the same time increasing the NSS score by a significant margin.
This result outline that AUC is in line with recent arguments by \cite{bylinskii2016different} and the NSS score is a better evaluation metric for saliency datasets based on fixation patterns.
The reason for the large NSS loss is due to our models output has a high standard deviation, which lead to a pixel value decrease after normalisation when compute the NSS score. See Figure~\ref{fig:NSS}.
It would appear at first glance that the NSS score for our model prediction should be higher than the salient region map, as it covers more eye fixation locations---this is not the case. The  NSS score for the saliency map is 4.2324, for salient region map is 3.9484, and for our prediction is 3.1471. As the standard deviation for the saliency map is 0.0329, for salient region map is 0.0777, but for our model prediction is 0.1923, after normalisation, the max value for the saliency map is 11.6710, for salient region map is 12.7703, but for our model prediction is 4.9508, which will lead to a large decline in the NSS computation.
%The Figure~\ref{fig:results} is some result examples of our models and other state-of-the-art for which the code is publicly available:
%\begin{figure}[h!]
%	\centering
%	\subfigure[image 1]{\includegraphics[width=20mm]{1}}
%	\subfigure[GT]{\includegraphics[width=20mm]{1_s}}
%	\subfigure[ours]{\includegraphics[width=20mm]{1_3}}
%	\subfigure[eDN]{\includegraphics[width=20mm]{edn1}}
%	\subfigure[CovSal]{\includegraphics[width=20mm]{cos1}}
%	\subfigure[GBVS]{\includegraphics[width=20mm]{gbv1}}
%	\subfigure[image 2]{\includegraphics[width=20mm]{2}}
%	\subfigure[GT]{\includegraphics[width=20mm]{2_s}}
%	\subfigure[ours]{\includegraphics[width=20mm]{2_3}}
%	\subfigure[eDN]{\includegraphics[width=20mm]{edn2}}
%	\subfigure[CovSal]{\includegraphics[width=20mm]{cos2}}
%	\subfigure[GBVS]{\includegraphics[width=20mm]{gbv2}}
%	\subfigure[image 3]{\includegraphics[width=20mm]{3}}
%	\subfigure[GT]{\includegraphics[width=20mm]{3_s}}
%	\subfigure[ours]{\includegraphics[width=20mm]{3_3}}
%	\subfigure[eDN]{\includegraphics[width=20mm]{edn3}}
%	\subfigure[CovSal]{\includegraphics[width=20mm]{cos3}}
%	\subfigure[GBVS]{\includegraphics[width=20mm]{gbv3}}
%	\subfigure[image 4]{\includegraphics[width=20mm]{4}}
%	\subfigure[GT]{\includegraphics[width=20mm]{4_s}}
%	\subfigure[ours]{\includegraphics[width=20mm]{4_3}}
%	\subfigure[CovSal]{\includegraphics[width=20mm]{cos4}}
%	\subfigure[GBVS]{\includegraphics[width=20mm]{gbv4}}
%	\caption{Example results comparison, here we use the saliency map as the GT}
%	\label{fig:results}
%\end{figure}\\
\subsection{Reformulation Gain}
\begin{figure}[h!]
	\centering
	\begin{subfigure}{0.23\textwidth}
		\centering
		\includegraphics[width=\textwidth]{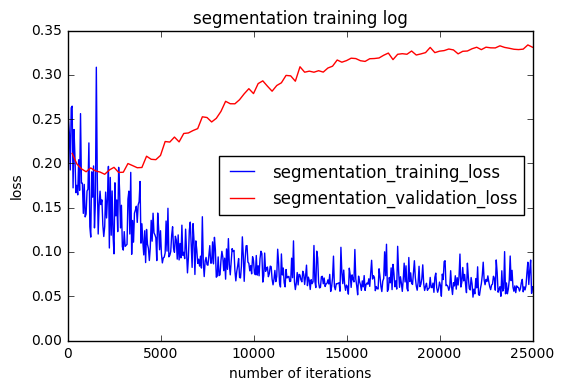}
		\caption{segmentation training log}
		\label{fig:seg}
	\end{subfigure}
	\hfill
	\centering
	\begin{subfigure}{0.23\textwidth}
		\centering
		\includegraphics[width=\textwidth]{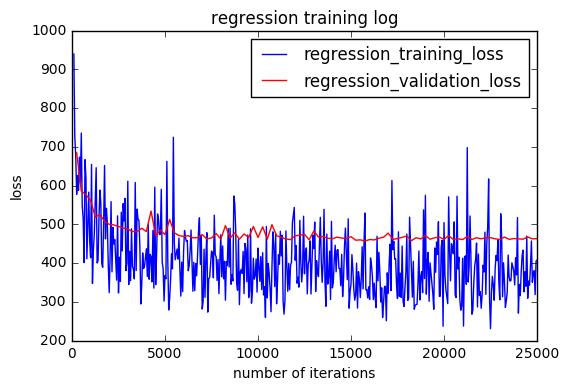}
		\caption{regression training log}
		\label{fig:reg}
	\end{subfigure}
	\hfill
	\centering
	\begin{subfigure}{0.23\textwidth}
		\centering
		\includegraphics[width=\textwidth]{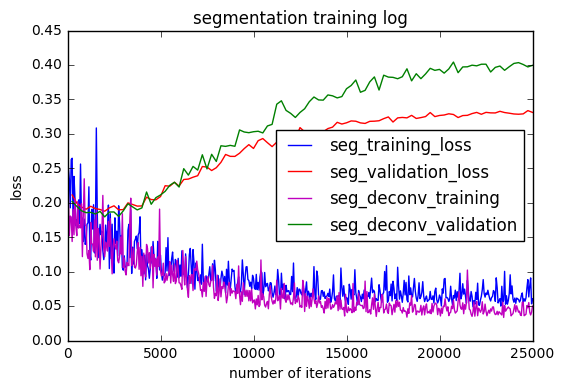}
		\caption{unpooling and deconvolution in segmentation}
		\label{fig:seg_un_de}
	\end{subfigure}
	\hfill
	\centering
	\begin{subfigure}{0.23\textwidth}
		\centering
		\includegraphics[width=\textwidth]{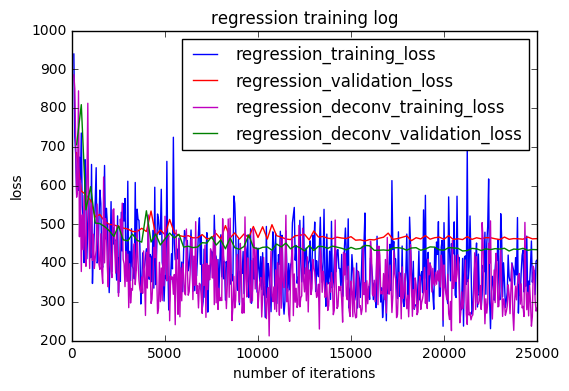}
		\caption{unpooling and deconvolution in regression}
		\label{fig:reg_un_de}
	\end{subfigure}
	\caption{Comparison of segmentation versus regression and unpooling versus deconvolution.}
	\label{fig:conv}
\end{figure}

One important advantage of reformulating the problem as semantic segmentation is that it is a simpler problem to learn. 
To demonstrate this we compare the training convergence of the our segmentation model with a regression model using the same architecture, same dataset and same learning strategy, but treating saliency map as the ground truth and using a common Euclidean loss. The results are illustrated in Figure~\ref{fig:conv}.

In Figures~\ref{fig:seg} and~\ref{fig:reg}, we can see that the proposed method has a much faster convergence speed: it only requires 2,000 iterations to converge (only two epochs). In contrast, the regression-based model is slow to converge and oscillates easily. And in Figures~\ref{fig:seg_un_de} and~\ref{fig:reg_un_de}, we can see that using the deconvolution method in the decoder performs better for both segmentation and regression.

\subsection{General Features Learned by The Network}
In recent years, several model for saliency prediction were proposed using deep learning method, but none of them visualised \textit{what} is learned by those models. Therefore, we adopt the popular deep neural network visualisation technique in \cite{zeiler2014visualizing} to analyse our model. There are three main processes in the visualisation: upsampling, deconvolution, and non-linearity 
(usually a ReLU function)---the reverse of the forward pass when input an image into the deep neural network. However, 
in this paper we are more interested in visualising to what patterns the deep neurons are attuned to rather than features for individual images. This required a modification of the upsampling process. The classic upsampling method in feature visualisation is \textit{unpooling}, using the pooling indices in the forward pass to do unpooling. Because pooling indices only exist when processing an actual image through the network, these indices are not available when visualising a neuron's receptive field in abstraction from any input. Hence, in order to visualise general individual neuron's receptive fields, we set the pooled feature map as a sparse matrix (with only one non-zero value) and do upsampling by repeating this sparse matrix. Here, we show the visualisation results of our model: the features learned by the last layer of the encoder part in our network, there are 512 neurons. Here, we show the general feature for the first 64 neurons.
\begin{figure}%[h!]
	\centering
	\includegraphics[width=\columnwidth]{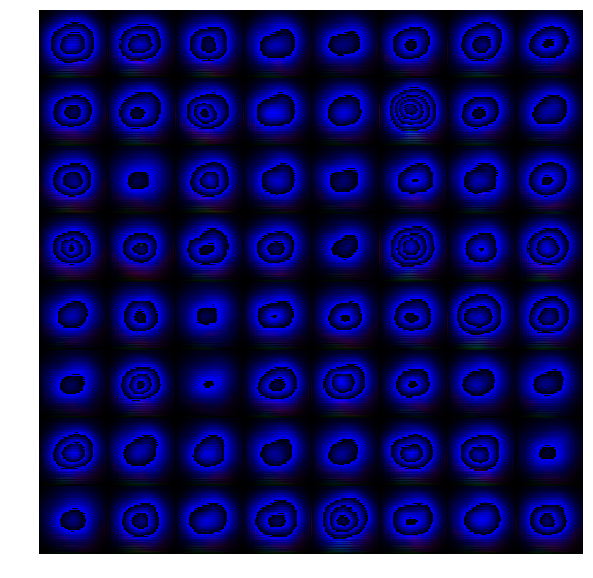}
	\caption{Illustration of the receptive field of the features learnt by the first 64 neurons of top encoder layer in our architecture.}
	\label{fig:general_feature}
\end{figure}
From the Figure~\ref{fig:general_feature}, we can see that the features learned by our network are very similar to central-surround patterns, which is consistent with the research in Psychology \cite{desimone1995neural}.
\section{Conclusion}
In this work, we propose to reformulate the saliency prediction problem as an image segmentation problem, according to different saliency levels rather than the traditional pix-wise saliency value prediction. We adopt the encoder-decoder architecture in semantic segmentation to do salient region segmentation. Our results show that even if we treat it as a segmentation problem, it can still challenge the state-of-the-art performance, and the proposed model trains faster and more reliably than regression-based models. Finally, we also demonstrated that the learnt deep features are consistent with the centre-surround hypothesis.

\bibliographystyle{ieeetr}
\bibliography{all}

\begin{thebibliography}{10}

\bibitem{Borji2013b}
A.~Borji and L.~Itti, ``{State-of-the-srt in Visual Attention Modeling},'' {\em
  IEEE Trans. on Pattern Analysis and Machine Intelligence (TPAMI)}, vol.~35,
  no.~1, pp.~185--207, 2013.

\bibitem{Treisman1980}
A.~M. Treisman and G.~Gelade, ``{A feature-integration theory of attention.},''
  {\em Cognitive psychology}, vol.~12, no.~1, pp.~97--136, 1980.

\bibitem{Itti2001}
L.~Itti and C.~Koch, ``{Computational modelling of visual attention.},'' {\em
  Nature reviews. Neuroscience}, vol.~2, no.~3, pp.~194--203, 2001.

\bibitem{kummerer2014deep}
M.~K{\"u}mmerer, L.~Theis, and M.~Bethge, ``Deep gaze i: Boosting saliency
  prediction with feature maps trained on imagenet,'' {\em arXiv preprint
  arXiv:1411.1045}, 2014.

\bibitem{Kummerer2016}
M.~K{\"{u}}mmerer, T.~S.~A. Wallis, and M.~Bethge, ``{DeepGaze II: Reading
  fixations from deep features trained on object recognition},'' {\em arXiv
  preprint}, pp.~1--16, 2016.

\bibitem{Jetley2016a}
S.~Jetley, N.~Murray, and E.~Vig, ``{End-to-End Saliency Mapping via
  Probability Distribution Prediction},'' {\em Proceedings of the IEEE
  Conference on Computer Vision and Pattern Recognition}, pp.~5753--5761, 2016.

\bibitem{kruthiventi2015deepfix}
S.~S. Kruthiventi, K.~Ayush, and R.~V. Babu, ``Deepfix: A fully convolutional
  neural network for predicting human eye fixations,'' {\em arXiv preprint
  arXiv:1510.02927}, 2015.

\bibitem{Itti2000a}
L.~Itti and C.~Koch, ``{A saliency-based search mechanism for overt and covert
  hifts of visual attention},'' {\em Vision Research}, vol.~40, no.~10-12,
  pp.~1489--1506, 2000.

\bibitem{Ouerhani2000}
N.~Ouerhani and H.~Hugli, ``{Computing visual attention from scene depth},''
  {\em Proceedings 15th International Conference on Pattern Recognition.
  ICPR-2000}, vol.~1, pp.~375--378, 2000.

\bibitem{Kienzle2009a}
W.~Kienzle, M.~O. Franz, B.~Sch{\"{o}}lkopf, and F.~A. Wichmann,
  ``{Center-surround patterns emerge as optimal predictors for human saccade
  targets},'' {\em Journal of vision}, vol.~9(5), no.~7, pp.~1--15, 2009.

\bibitem{vig2014large}
E.~Vig, M.~Dorr, and D.~Cox, ``Large-scale optimization of hierarchical
  features for saliency prediction in natural images,'' in {\em Proceedings of
  the IEEE Conference on Computer Vision and Pattern Recognition},
  pp.~2798--2805, 2014.

\bibitem{Kruthiventi2015}
S.~S.~S. Kruthiventi, K.~Ayush, and R.~V. Babu, ``{DeepFix: A Fully
  Convolutional Neural Network for predicting Human Eye Fixations},'' {\em
  arXiv Preprint}, pp.~1--11, 2015.

\bibitem{Corniaa}
M.~Cornia, ``{Predicting Human Eye Fixations via an LSTM-based Saliency
  Attentive Model},'' pp.~1--10.

\bibitem{Long2015}
J.~Long, E.~Shelhamer, and T.~Darrell, ``{Fully Convolutional Networks for
  Semantic Segmentation ppt},'' {\em Proceedings of the IEEE Conference on
  Computer Vision and Pattern Recognition}, pp.~3431--3440, 2015.

\bibitem{Badrinarayanan2015}
V.~Badrinarayanan, A.~Kendall, and R.~Cipolla, ``{SegNet: A Deep Convolutional
  Encoder-Decoder Architecture for Image Segmentation},'' {\em CVPR 2015},
  p.~5, 2015.

\bibitem{zeiler2014visualizing}
M.~D. Zeiler and R.~Fergus, ``Visualizing and understanding convolutional
  networks,'' in {\em European conference on computer vision}, pp.~818--833,
  Springer, 2014.

\bibitem{Eigen2014}
D.~Eigen and R.~Fergus, ``{Predicting Depth, Surface Normals and Semantic
  Labels with a Common Multi-Scale Convolutional Architecture},'' 2014.

\bibitem{simonyan2014very}
K.~Simonyan and A.~Zisserman, ``Very deep convolutional networks for
  large-scale image recognition,'' {\em arXiv preprint arXiv:1409.1556}, 2014.

\bibitem{Jia2014}
Y.~Jia, E.~Shelhamer, J.~Donahue, S.~Karayev, J.~Long, R.~Girshick,
  S.~Guadarrama, and T.~Darrell, ``{Caffe: Convolutional Architecture for Fast
  Feature Embedding},'' {\em ACM International Conference on Multimedia},
  pp.~675--678, 2014.

\bibitem{jiang2015salicon}
M.~Jiang, S.~Huang, J.~Duan, and Q.~Zhao, ``Salicon: Saliency in context,'' in
  {\em Proceedings of the IEEE conference on computer vision and pattern
  recognition}, pp.~1072--1080, 2015.

\bibitem{judd2009learning}
T.~Judd, K.~Ehinger, F.~Durand, and A.~Torralba, ``Learning to predict where
  humans look,'' in {\em Computer Vision, 2009 IEEE 12th international
  conference on}, pp.~2106--2113, IEEE, 2009.

\bibitem{judd2012benchmark}
T.~Judd, F.~Durand, and A.~Torralba, ``A benchmark of computational models of
  saliency to predict human fixations,'' 2012.

\bibitem{Riche2013}
N.~Riche, M.~Duvinage, M.~Mancas, B.~Gosselin, and T.~Dutoit, ``{Saliency and
  human fixations: State-of-the-art and study of comparison metrics},'' {\em
  Proceedings of the IEEE International Conference on Computer Vision},
  pp.~1153--1160, 2013.

\bibitem{zhang2008sun}
L.~Zhang, M.~H. Tong, T.~K. Marks, H.~Shan, and G.~W. Cottrell, ``Sun: A
  bayesian framework for saliency using natural statistics,'' {\em Journal of
  vision}, vol.~8, no.~7, pp.~32--32, 2008.

\bibitem{mit-saliency-benchmark}
Z.~Bylinskii, T.~Judd, A.~Borji, L.~Itti, F.~Durand, A.~Oliva, and A.~Torralba,
  ``{MIT} saliency benchmark.''

\bibitem{bylinskii2016different}
Z.~Bylinskii, T.~Judd, A.~Oliva, A.~Torralba, and F.~Durand, ``What do
  different evaluation metrics tell us about saliency models?,'' {\em arXiv
  preprint arXiv:1604.03605}, 2016.

\bibitem{desimone1995neural}
R.~Desimone and J.~Duncan, ``Neural mechanisms of selective visual attention,''
  {\em Annual review of neuroscience}, vol.~18, no.~1, pp.~193--222, 1995.

\end{thebibliography}

\end{document}